\documentclass[letterpaper, 10 pt, conference]{ieeeconf}  

\IEEEoverridecommandlockouts  

\usepackage[table,xcdraw]{xcolor}

\usepackage{graphicx}
\usepackage{subfig}
\usepackage{booktabs}
\usepackage{multirow}
\usepackage{balance}

\usepackage{pifont}
\newcommand{\cmark}{\ding{51}}
\newcommand{\xmark}{\ding{55}}
\usepackage{svg}

\usepackage[nolist,nohyperlinks]{acronym}
\begin{acronym}
    \acro{IL}{imitation learning}
    \acro{DP}{Diffusion Policy}
    \acro{DoF}{degree of freedom} 
    \acroplural{DoF}[DoFs]{degrees of freedom} 
\end{acronym}

\usepackage{hyperref}
\hypersetup{
    hidelinks,
    colorlinks=true,
    linkcolor=blue,
    urlcolor=blue,
    citecolor=blue,
    allcolors=blue
    }
\usepackage{cite}

\title{\LARGE \bf
HannesImitation: Grasping with the Hannes Prosthetic Hand\\via Imitation Learning
}

\author{
Carlo Alessi\textsuperscript{1}, %
Federico Vasile\textsuperscript{1}, %
Federico Ceola\textsuperscript{1}, %
Giulia Pasquale\textsuperscript{1}, %
Nicolò Boccardo\textsuperscript{2,3}, and %
Lorenzo Natale\textsuperscript{1}
\thanks{\textsuperscript{1} Humanoid Sensing and Perception, Istituto Italiano di Tecnologia, 16163 Genoa, Italy. ({\tt\small carlo.alessi@iit.it})}%
\thanks{\textsuperscript{2} Rehab Technologies Lab, Istituto Italiano di Tecnologia, 16163 Genoa, Italy.}%
\thanks{\textsuperscript{3} Open University Affiliated Research Center at Istituto Italiano di Tecnologia (ARC@IIT), Genova, Italy. The ARC@IIT is part of the Open University, Milton Keynes MK7 6AA, United Kingdom.}%
}

\begin{document}

\maketitle
\thispagestyle{empty}
\pagestyle{empty}

\begin{abstract}
Recent advancements in control of prosthetic hands have focused on increasing autonomy through the use of cameras and other sensory inputs. These systems aim to reduce the cognitive load on the user by automatically controlling certain degrees of freedom. In robotics, imitation learning has emerged as a promising approach for learning grasping and complex manipulation tasks while simplifying data collection. Its application to the control of prosthetic hands remains, however, largely unexplored. Bridging this gap could enhance dexterity restoration and enable prosthetic devices to operate in more unconstrained scenarios, where tasks are learned from demonstrations rather than relying on manually annotated sequences. To this end, we present \textit{HannesImitationPolicy}, an imitation learning-based method to control the Hannes prosthetic hand, enabling object grasping in unstructured environments. Moreover, we introduce the \textit{HannesImitationDataset} comprising grasping demonstrations in table, shelf, and human-to-prosthesis handover scenarios. We leverage such data to train a single diffusion policy and deploy it on the prosthetic hand to predict the wrist orientation and hand closure for grasping. Experimental evaluation demonstrates successful grasps across diverse objects and conditions. Finally, we show that the policy outperforms a segmentation-based visual servo controller in unstructured scenarios. Additional material is provided on our project page: \href{https://hsp-iit.github.io/HannesImitation/}{https://hsp-iit.github.io/HannesImitation}.
\end{abstract}

\section{Introduction}
Learning-based robot control is emerging as a dominant paradigm for solving complex grasping and manipulation tasks across diverse robotic platforms \cite{ibarz2021train, falotico2024learning}. In parallel, upper limb prostheses have evolved into sophisticated robotic devices with multiple \acp{DoF} \cite{trent2020narrative}. These advancements introduce new challenges and opportunities for learning-based methods in prosthetics.

\begin{figure}[tp]
    \centering
    \includegraphics[width=1\columnwidth]{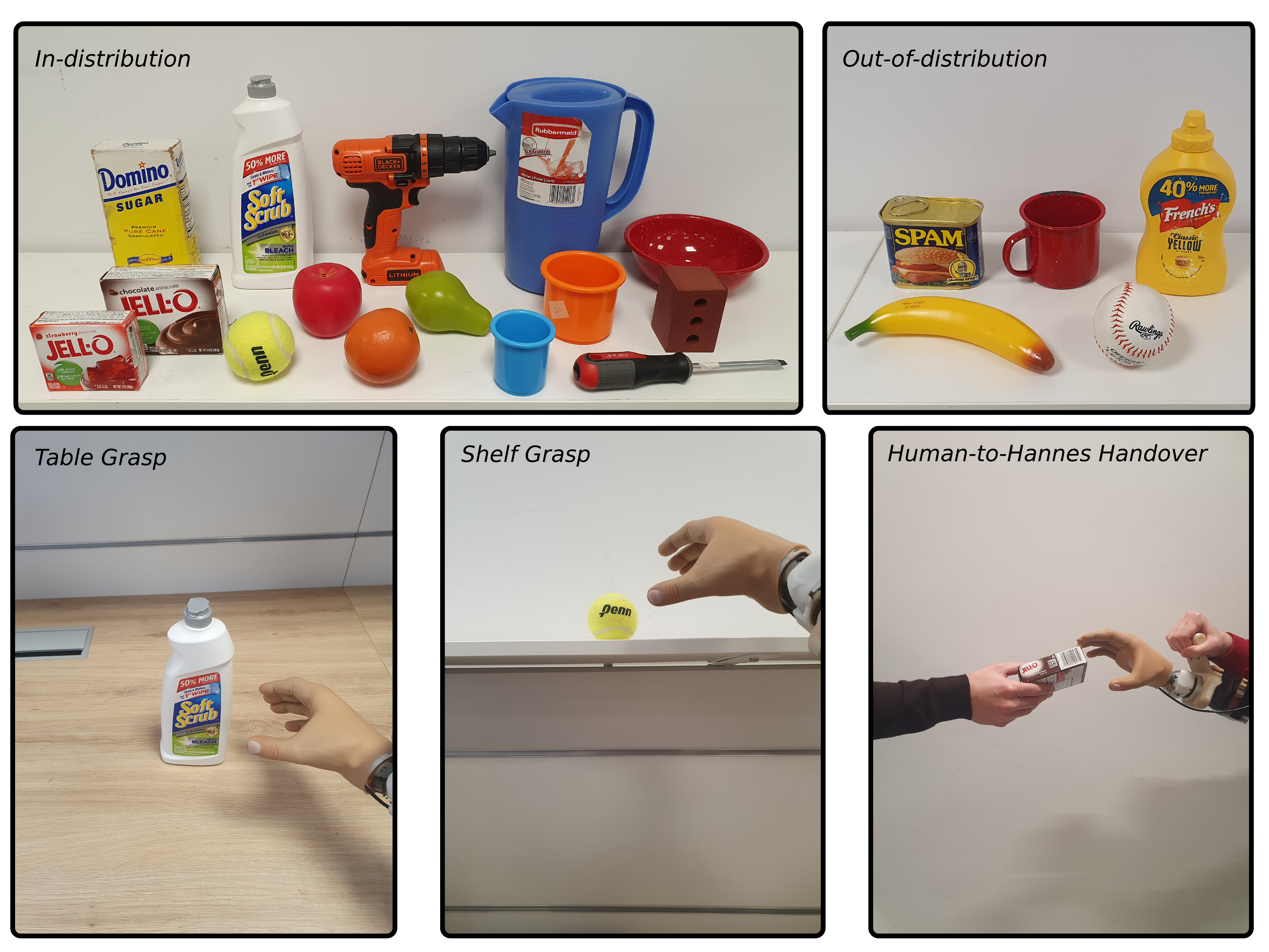}
    \caption{We propose \textit{HannesImitation}, an imitation learning-based approach that trains a single grasping policy across diverse objects and environments. The learned policy is deployed on the Hannes hand~\cite{laffranchi2020hannes}, enabling control of wrist orientation and fingers closure from an eye-in-hand camera.}
    \label{fig:overview}
\end{figure}

Most commercial prostheses leverage electromyography (EMG) or mechanomyography (MMG) signals from residual muscle activity to decode intended movements~\cite{oskoei2007myoelectric, chen2023review}. Generally, two surface EMG electrodes are placed on antagonist muscles to decode intended movements.
However, this approach limits intuitive control and dexterity, as users can typically control only one joint at a time. As the number of \acp{DoF} increases, dexterous manipulation becomes more challenging, leading to higher cognitive load~\cite{amsuess2014extending}. This limitation contributes to user dissatisfaction and, ultimately, device abandonment~\cite{biddiss2007upper}. Other strategies include threshold-based, proportional control and pattern recognition methods~\cite{di2021hannes} or incremental learning~\cite{egle_preliminary_2023}. Conversely, machine learning models to predict hand motions were investigated to reduce the cognitive burden and enable precise myocontrol~\cite{scheme2011electromyogram}.
Despite the effectiveness of these techniques in laboratory settings, pattern recognition remains unstable due to the non-stationarity of EMG signals caused by muscle fatigue, electrode displacement and difference in arm posture~\cite{kyranou2018causes}.
Leveraging alternative input modalities, such as images, presents a viable solution to these challenges~\cite{marinelli2023active}. Specifically, additional input sources can either complement the user commands or be utilized by a semi-autonomous system to execute certain stages of the grasping action, aligning with the \textit{shared-autonomy} framework~\cite{zhuang2019shared, gardner2020multimodal, guo2023toward}. However, these approaches typically learn from labeled data, that are difficult to acquire. Instead, we believe that learning a policy from demonstrations could unlock new possibilities. First, demonstrations inherently encapsulate all the necessary actions to accomplish a task, allowing semi-autonomous control of both wrist orientation---similar to prior work~\cite{starke2022semi,castro2022continuous,vasile2025continuous}---and hand closure,
which could potentially reduce the cognitive burden on the user. Second, this approach facilitates deployment in tasks which require sophisticated labels. For instance, consider a human-to-prosthesis handover for a mug: if the human holds the mug by the handle, the prosthesis should grasp it from the body. Manually specifying this contextual information, e.g., with affordance segmentation masks or grasping poses, is impractical. Instead, learning from demonstration (also referred to as behavior cloning in the robotics literature) allows to autonomously learn from sequences of paired action-images demonstrating the task.

This paper introduces a novel \ac{IL}-based control pipeline for prosthetic hands equipped with a camera embedded into the palm. By leveraging the generative capabilities of diffusion models, the proposed approach enables robust grasping across diverse objects and scenarios (Fig.~\ref{fig:overview}). The paper is organized as follows. First, we review related work on \ac{IL} for robotics and vision-based prosthetic control, highlighting the potential advantages in applying \ac{IL}-based methods to control prosthetic devices (Sec.~\ref{sec:related_works}). Next, we introduce the \textit{HannesImitationDataset} and describe how we adapt \ac{DP}~\cite{chi2023diffusionpolicy} to achieve high-frequency grasping with the Hannes prosthesis (Sec.~\ref{sec:materials_and_methods}). We release the dataset to foster further developments on \ac{IL} for prosthetics. In Sec.~\ref{sec:results_and_discussion}, we present an extensive experimental validation, offering insights into learning-based prosthetic grasping from demonstrations. Finally, we discuss future research directions and summarize our key findings (Sec.~\ref{sec:conclusion}).
In summary, the major contributions of \textit{HannesImitation} are:

\begin{enumerate}
    \item \textbf{HannesImitationPolicy}. 
    A novel DP-based approach for prosthetic grasping across a variety of objects and environments. The learned policy is deployed on the Hannes prosthesis to control wrist orientation and hand closure from visual input.

    \item \textbf{HannesImitationDataset}. 
    A collection of grasping demonstrations using the Hannes hand in both structured and unstructured environments. To the best of our knowledge, this is the first dataset for \ac{IL} with prosthetic hands.
\end{enumerate}

\section{Related Work}
\label{sec:related_works}

\subsection{Robot Manipulation with Behavior Cloning}

The latest vision-language-conditioned \ac{IL} models have shown promising generalization capabilities over different manipulation tasks~\cite{zitkovich2023rt, oxe, kim2024openvla, black2024pi_0}. Conversely, diffusion-based approaches like \ac{DP}~\cite{chi2023diffusionpolicy} and its Flow Matching~\cite{zhang2024affordance} variant achieve state-of-the-art performance for single-task learning.
DP~\cite{chi2023diffusionpolicy} is a widely adopted \ac{IL} method for robotic manipulation tasks. Its policy formulation as a Denoising Diffusion Probabilistic Model (DDPM)~\cite{ho2020denoising} brings several advantages, including handling multi-modal actions, scalability to high-dimensional action spaces, and training stability, outperforming classical regression models for action prediction~\cite{zhang2018deep}.
While behavior cloning is an efficient method to map robot observations to actions, data collection for policy learning could be time-consuming as it often requires teleoperation and specific hardware~\cite{oxe}. Some works overcome this limitation by simplifying data collection using hand-held grippers~\cite{shafiullah2023bringing, chi2024universal} and transfer policies trained on those data zero-shot on different robots.

In this work, we adapt DP to prosthetic control, specifically to solve grasping and human-to-prosthesis handover tasks with the Hannes prosthesis, leveraging an \textit{eye-in-hand} camera and proprioception with an approach similar to~\cite{chi2024universal}.

\subsection{Vision-based Prosthetic Control}

Early vision-based prosthetic systems employed image-processing techniques to estimate the object size and distance, automatically determining the most appropriate hand aperture and grasp type~\cite{dovsen2010cognitive, markovic2015sensor}. More recent approaches rely on advanced computer vision algorithms---such as object detection and segmentation~\cite{cirelli2023semiautonomous, 10235885}, grasp type prediction~\cite{vasile2022grasp, 10731261, zandigohar2019towards} and object mesh estimation~\cite{10375204}---to enrich information extraction and enable more precise and dexterous prosthetic control.

A key strategy for applying these techniques in practical prosthetic scenarios is through the \textit{shared-autonomy} framework~\cite{gardner2020multimodal, 10643668}, where additional input sources (e.g., images, tactile feedback, IMUs) or alternative viewpoints (e.g., gaze tracking~\cite{zandigohar2024multimodal}) assist the user in executing grasping tasks. For instance, in~\cite{starke2022semi}, the user can activate the automatic system via EMG signals and aim at the object to trigger a grasping suggestion using the \textit{eye-in-hand} camera. If needed, they can modify the proposed grasp using an IMU before executing the final selection on the device. Other approaches also employ \textit{eye-in-hand} cameras to predict the grasping object part as the hand approaches~\cite{vasile2025continuous, stracquadanio2025bring, zandigohar2019towards}. Similarly, \cite{castro2022continuous} introduces a proportional controller based on visual input to continuously adjust the prosthesis configuration before grasping. These methods share a common principle: the user retains responsibility for either initiating or finalizing the grasp by selecting the target object or commanding hand closure. While this maintains user involvement, it hinders the natural flow of the approach-to-grasp sequence.

To overcome this limitation, in this work, we leverage recent advances in learning-based algorithms from demonstrations to collect natural approaching sequences and deploy the system on the Hannes hand.

\section{Materials \& Methods}
\label{sec:materials_and_methods}

\subsection{The Hannes Prosthetic Device}
We test the proposed methods on the Hannes hand~\cite{laffranchi2020hannes}, considering the setup for a right arm trans-radial amputation.
In \cite{boccardo2023development}, Hannes has been extended to three \acp{DoF}: wrist flexion/extension (\textbf{Wrist F/E}), wrist pronation/supination (\textbf{Wrist P/S}) and fingers opening/closing (\textbf{Hand O/C}). The \textbf{Wrist~F/E} and \textbf{Wrist P/S} are revolute joints that are orthogonal and intersect at a common point. The \textbf{Hand O/C} is a single \ac{DoF} being the fingers actuated all together using one motor. 
The \textbf{Hand O/C} and \textbf{Wrist F/E} joints are equipped with position encoders and are controlled in position. In contrast, the \textbf{Wrist P/S} is controlled using velocity inputs, as it operates without an integrated position encoder. Finally, a tiny RGB camera is embedded into the palm of the prosthesis to enable visual feedback\footnote{The data collection and policy deployment experiments were conducted in line with the Declaration of Helsinki and approved by the local ethical committee (CER Liguria Ref. 11554 of October 18, 2021).}.

\begin{table}[tp]
\vspace{2mm}
\centering
\caption{Overview of the proposed \textit{HannesImitationDataset} for grasping and human-to-prosthesis handover tasks collected with the Hannes prosthesis. Objects and scenarios are shown in Fig.~\ref{fig:overview}.}
\label{tab:hannes_imitation_dataset}
\resizebox{1\columnwidth}{!}{%
\begin{tabular}{@{}lccc@{}}
\toprule
\textbf{Scenario / Task (\textbf{\#})}                           & \textbf{Clutter} & \textbf{Objects}       & \textbf{\#Demo/object} \\ \midrule
\textit{Table Grasp} (\textbf{\#1})    & \xmark & 15 YCB   & 10    \\
\midrule
\textit{Shelf Grasp} (\textbf{\#2})    & \xmark      & 15 YCB   & 10     \\
\midrule
\textit{Human-to-Hannes Handover} (\textbf{\#3}) & \cmark   & 15 YCB      & 10 \\
\bottomrule
\end{tabular}
} 
\end{table}

\subsection{HannesImitationDataset}
We present the \textit{HannesImitationDataset} for learning control policies with the Hannes prosthetic hand \cite{laffranchi2020hannes} via behavior cloning. The dataset comprises three collections of grasping and human-to-prosthesis handover tasks performed in three different unstructured scenarios:

\begin{itemize}
    \item \textit{Table Grasp} (\textbf{\#1}). The user drives the prosthesis to grasp objects from a table with a wooden-style pattern surface.
    \item \textit{Shelf Grasp} (\textbf{\#2}). The user guides the prosthetic hand to grasp objects from the top of a white shelf. This scenario introduces a different visual perspective and requires distinct wrist and hand movements compared to \textbf{\#1}, challenging the policy to adapt to different grasping angles and spatial constraints.
    \item \textit{Human-to-Hannes Handover} (\textbf{\#3}). A subject hands an object over to the prosthetic hand controlled by the user. This scenario is particularly challenging due to its unstructured environment, potential object occlusions and background.
\end{itemize}

We remark that the handover experiment is conducted to assess the capabilities of the grasping policy when trained on both common standard scenarios (e.g., tabletop) and more unconstrained conditions. The subject did not receive additional instructions and acted naturally. 
The goal is to assess how well the policy learns when exposed to diverse conditions, testing its ability to handle variations in object appearance or positioning.

The \textit{HannesImitationDataset} comprises the Hannes prosthesis grasping 15 objects from the YCB dataset \cite{calli2015benchmarking}. 
Thanks to the different physical features like shape, mass, and color of the considered objects, \textit{HannesImitationDataset} contains a heterogeneous set of grasping demonstrations. We ensure grasp variability by collecting 10 demonstrations for each object and scenario, randomizing the initial pose of the object, and varying the approach velocity. We also collect data starting from different wrist flexion/extension and pronation/supination joint positions to mimic conditions where a user grasps an object with the prosthesis in a random configuration. In total, the \textit{HannesImitationDataset} comprises 450 demonstrations, and Tab.~\ref{tab:hannes_imitation_dataset} provides an overview of its key characteristics.

To collect the demonstrations, we continuously control the fingers and wrist joints of the prosthesis using a keyboard interface. Each demonstration records: (i) encoder measurements for hand opening/closing and wrist flexion/extension, (ii) visual observations from the \textit{eye-in-hand} camera, and (iii) the three control actions for the \acp{DoF} of the policy:

\begin{itemize}
    \item \textbf{Hand O/C} controls the position of the hand opening or closing, ranging from 0 units (fully open) to 100 units (fully closed).
    \item \textbf{Wrist F/E} controls the position of the wrist flexion or extension, ranging from 0 units (full flexion) to 100 units (full extension).
    \item \textbf{Wrist P/S} controls the velocity of the wrist pronation or supination, ranging from -30 (outward rotation) to +30 (inward rotation).
\end{itemize}

We use the \textit{HannesImitationDataset} to train and validate the \textit{HannesImitationPolicy} to control the Hannes prosthesis with action sequences generated by DP, starting from encoder measurements and camera frames.

\begin{figure}[tp]
    \vspace{2mm}
    \centering
    \includegraphics[width=1\columnwidth]{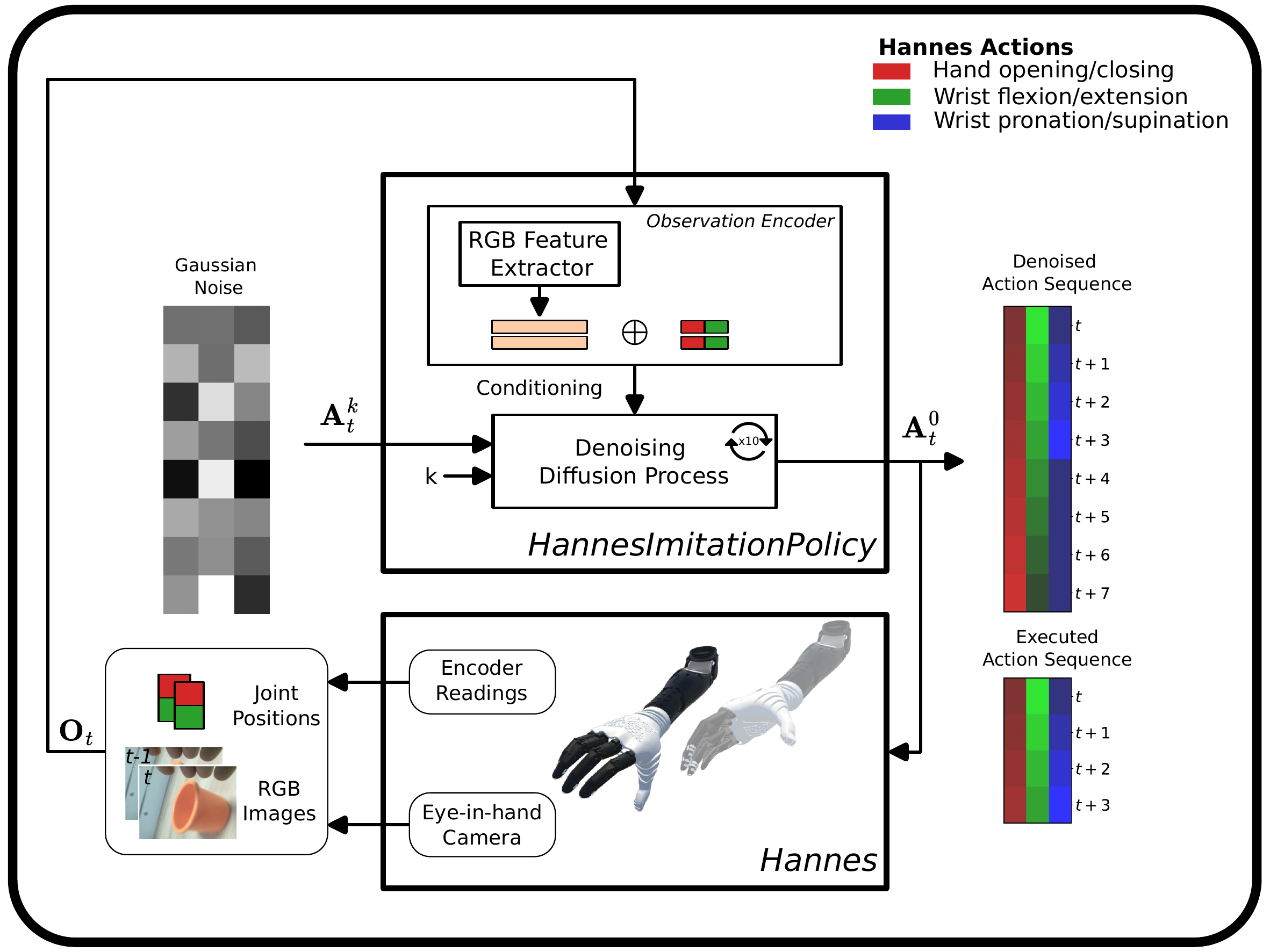}
    \caption{Control architecture of \textit{HannesImitationPolicy}.}
    \label{fig:hannes_imitation_scheme}
\end{figure}

\subsection{HannesImitationPolicy} 
\textit{HannesImitationPolicy} is a pipeline to control the hand and wrist of the Hannes prosthesis for grasping objects with \ac{DP}. \ac{DP} is a recent \ac{IL} method to generate robot behaviors by representing visuomotor policies as DDPMs. We adapt DP to grasp objects with Hannes, enabling high-frequency control from visual and proprioceptive data acquired from the prosthesis.

\subsubsection{Control architecture}
Fig.~\ref{fig:hannes_imitation_scheme} shows the proposed control architecture. We leverage two different sources of information from the Hannes prosthesis as observations $\mathbf{O}_t$ for DP: (i) sensor measurements of hand opening/closing and the wrist flexion/extension positions, and (ii) RGB images from the \textit{eye-in-hand} camera mounted on the palm. 
We extract features from images with a ResNet-18~\cite{he2016deep} trained from scratch.
We then concatenate encoder measurements and the visual feature vectors at a given timestep $t$ to obtain the input for the \textit{Denoising Diffusion Process}. 
At each time step $t$, the policy takes as input the latest two observations and predicts an action sequence $\mathbf{A}_t$ of eight steps. $\mathbf{A}_t$ controls hand opening/closing (\textbf{Hand O/C}), wrist flexion/extension (\textbf{Wrist F/E}), and wrist pronation/supination (\textbf{Wrist P/S}). To balance smooth long-horizon planning and prompt reaction required by prosthetic control, we execute a shorter action sequence of four steps before replanning.
As in DP, we employ the 1D temporal convolutional U-Net architecture as noise prediction network $\epsilon_\mathbf{\theta}$. However, to optimize computation and inference time while maintaining satisfactory performance, we reduce the number of parameters of the U-Net. We use two convolutional layers of size 32 and 64, and we set the kernel size to 3. Additionally, we reduce the dimension of the diffusion step embedding to 32.

\begin{figure}[t]
\vspace{2mm}
    \centering
    \includegraphics[width=1\columnwidth]{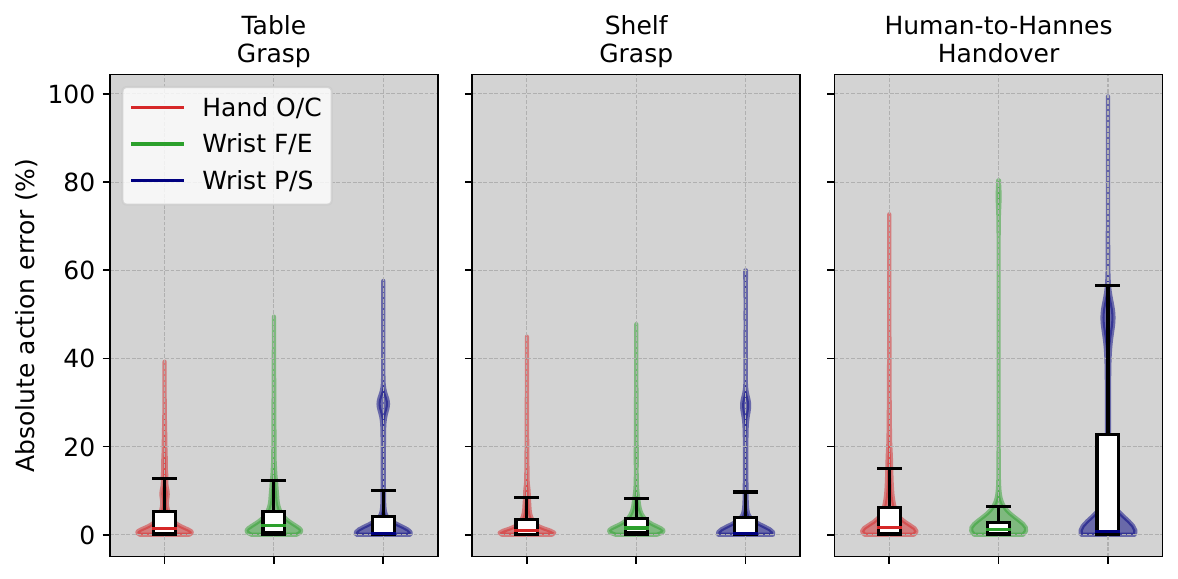}
    \caption{Absolute action error distributions for the \textit{HannesImitationPolicy} on the validation set, separated by hand and wrist motions across three tasks. 
    }
    \label{fig:error_distributions_validation_set}
\end{figure}

\subsubsection{Policy training}
\ac{DP} training is composed of two steps. First, given ground-truth action sequences $\mathbf{A}^0$, it selects a denoising iteration $k$, samples random gaussian noise $\epsilon^k$ with the appropriate variance for iteration $k$, and perturbs the original action samples as $\mathbf{A}^k = \mathbf{A}^0 + \epsilon^k$. Then, \ac{DP} trains a denoising network $\epsilon_\mathbf{\theta}$ with parameters $\mathbf{\theta}$ to predict the noise added to the data, optimizing the following objective:

\begin{equation}
\mathcal{L} = MSE(\epsilon^k, \epsilon_{\theta}(\mathbf{O}, \mathbf{A}^k, k)).
\end{equation}

\noindent This loss encourages the network to recover clean action sequences from different corruption levels.
We train \ac{DP} on the \textit{HannesImitationDataset}, splitting training and validation sets with a $0.8/0.2$ ratio. We train \ac{DP} with denoising iterations $k \in [1,10]$ for $100$ epochs using the AdamW optimizer with learning rate $1e{-}4$, weight decay $2e{-}4$, and batch size $256$.

\subsubsection{Policy inference}
\ac{DP} inference is modeled as a denoising process. This process first samples an initial action $\mathbf{A}_t^k$ from a standard gaussian distribution. Then DP iteratively denoises it using the output of the noise prediction network $\epsilon_\mathbf{\theta}(\mathbf{O}_t, \mathbf{A}_t^k, k)$. This process is repeated $k$ times to generate the denoised action sequence $\mathbf{A}_t^0$:

\begin{equation}
    \mathbf{A}_t^{k-1} = \alpha(\mathbf{A}_t^k - \gamma \epsilon_{\theta}(\mathbf{O}_t, \mathbf{A}_t^k, k) + \mathcal{N}(0, \sigma^2 I)).
\end{equation}

\noindent To comply with our aim to use the policy in adaptive prosthetic scenarios, which require high-frequency reactive control, we set the number of denoising iterations $k$ to 10.
The compact policy architecture and reduced number of diffusion steps allow an inference frequency of about 35 Hz on a standard laptop with an NVIDIA GeForce RTX 4060. This high control frequency ensures the responsiveness needed for seamless integration between the user driving the prosthesis and the grasping policy.

\section{Results \& Discussion}
\label{sec:results_and_discussion}

After training \textit{HannesImitationPolicy} as described in Sec.~\ref{sec:materials_and_methods}, we perform an offline analysis on the validation set of the \textit{HannesImitationDataset} (Sec.~\ref{sec:offline_validation}). Then, we conduct experiments to quantitatively analyze the performance of the policy when deployed on the physical Hannes hand (Sec.~\ref{sec:policy_deployment}). Finally, we evaluate the generalization capability of the policy by grasping unseen objects (Sec.~\ref{sec:generalization_unseen_objects}) and compare the \textit{HannesImitation} approach with a visual servo wrist controller (Sec.~\ref{sec:comparison_visual_servo}).

\subsection{Offline Validation of HannesImitationPolicy}
\label{sec:offline_validation}
We evaluate \textit{HannesImitationPolicy} offline on the 90 demonstrations of the validation set. For each demonstration, we compute absolute errors between ground-truth and predicted action sequences composed of four steps.
Fig.~\ref{fig:error_distributions_validation_set} shows the distributions of action error---grouped by scenario and action---normalized with respect to their action bounds. Since the errors are mainly concentrated around low values and exhibit similar trends, it suggests that the policy has learned to perform the tasks in different conditions. Within each scenario, the errors of the \textbf{Wrist P/S} tend to be slightly higher than the errors of the other two \acp{DoF}, likely due to the absence of a position encoder for that joint. Despite a limited number of outliers, the policy can still predict the wrist rotation conditioned on the \textit{eye-in-hand} observations. These results on the validation set supported the deployment of the \textit{HannesImitationPolicy} on Hannes.





\begin{table}[tp]
\vspace{2mm}
\caption{Test results of \textit{HannesImitationPolicy} deployed on the physical prosthesis. Task success rate on 3 scenarios for 15 YCB objects of the \textit{HannesImitationDataset}.}
\label{tab:test_results_YCB_train_1_4_7}
\resizebox{\columnwidth}{!}{%
\begin{tabular}{@{}l|c|c|c@{}}
\toprule
 &
  \multicolumn{1}{c}{\begin{tabular}[c]{@{}c@{}}\textbf{Table}\\ \textbf{Grasp}\end{tabular}} &
  \multicolumn{1}{|c}{\begin{tabular}[c]{@{}c@{}}\textbf{Shelf}\\ \textbf{Grasp}\end{tabular}} &
  \multicolumn{1}{|c}{\begin{tabular}[c]{@{}c@{}}\textbf{Human-to-Hannes}\\ \textbf{Handover}\end{tabular}} \\ 
  \midrule
\textbf{YCB Objects}      & 
\textbf{\#1} & 
\textbf{\#2} & 
\textbf{\#3} \\ 
\midrule
\texttt{004\_sugar\_box}               & $8/10$  &  $5/10$  &  $10/10$ \\
\texttt{008\_pudding\_box}  &   $10/10$  &  $6/10$   &   $9/10$  \\
\texttt{009\_gelatin\_box}            &  $10/10$   &  $7/10$   & $10/10$  \\
\texttt{013\_apple}                      &  $10/10$   & $6/10$    &  $10/10$  \\
\texttt{016\_pear}                      &  $6/10$   &  $10/10$   & $10/10$   \\
\texttt{017\_orange}                    &  $10/10$   &  $8/10$   &  $9/10$   \\
\texttt{019\_pitcher\_base}                    &  $7/10$   &  $9/10$   &  $8/10$  \\
\texttt{021\_bleach\_cleanser}            & $8/10$    &  $4/10$   &  $10/10$  \\ 
\texttt{024\_bowl}  &  $8/10$  & $10/10$ &  $8/10$ \\
\texttt{035\_power\_drill}               & $10/10$  &  $6/10$  &  $7/10$   \\
\texttt{044\_flat\_screwdriver}      &  $1/10$  &  $7/10$ &  $7/10$  \\
\texttt{056\_tennis\_ball}               &  $10/10$   & $9/10$    &  $10/10$   \\
\texttt{061\_foam\_brick}                   &  $9/10$   & $10/10$    &  $10/10$ \\
\texttt{065-b\_cups} (small blue)   &  $4/10$  & $3/10$ &   $9/10$ \\
\texttt{065-g\_cups} (big orange) & $10/10$  & $2/10$ & $7/10$ \\
\midrule
\multicolumn{1}{l}{\textbf{Aggregate success rate}} & 
\multicolumn{1}{c}{$80.6\%$} & 
\multicolumn{1}{c}{$68\%$} &
\multicolumn{1}{c}{$89.3\%$} \\
\midrule
\multicolumn{4}{l}{\textbf{Overall success rate:} $79.3\%$} \\
\bottomrule
\end{tabular}
}
\end{table}

\begin{figure*}[tp]
\vspace{2mm}
    \centering
    \subfloat[]{\includegraphics[width=1\columnwidth]{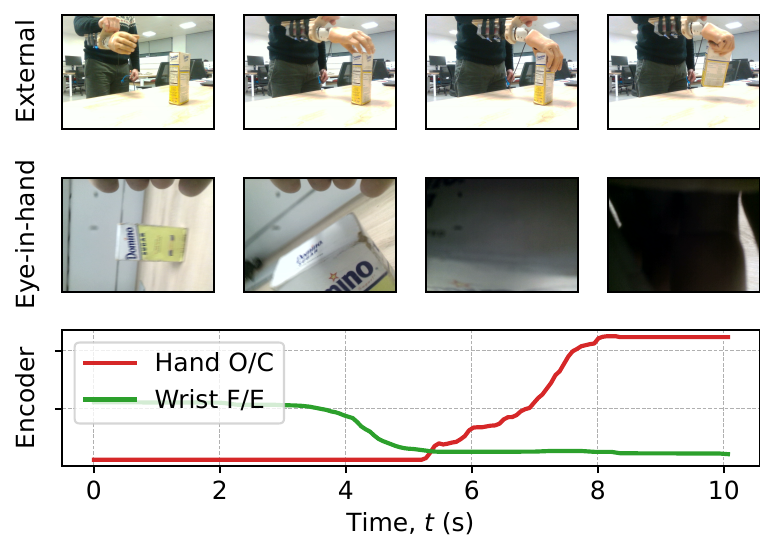}\label{fig:test_table_grasp_observations}}
    \subfloat[]{
    \includegraphics[width=1\columnwidth]{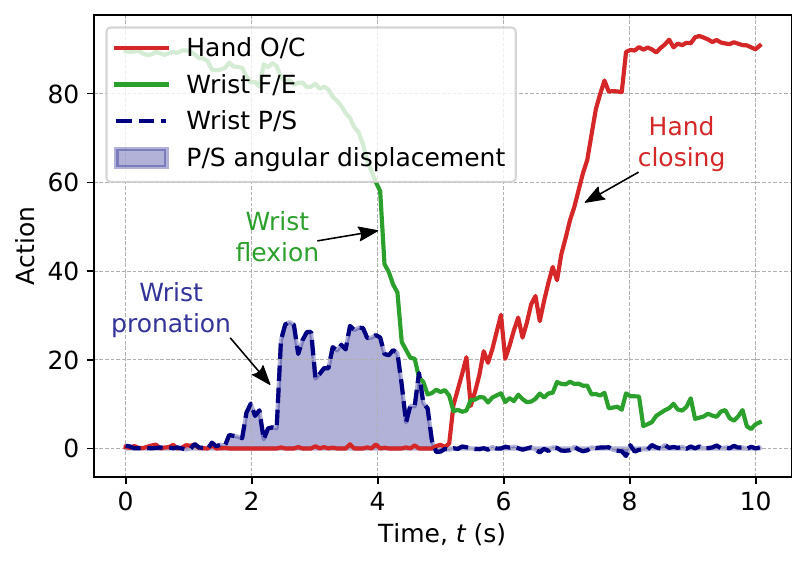}\label{fig:test_table_grasp_actions}}
    \caption{\textit{Table Grasp} (\textbf{\#1}). \textit{HannesImitationPolicy} deployed on the prosthetic hand to grasp the \texttt{004\_sugar\_box} object. (\textbf{a})~Top: External camera view showing the wrist motions during the approach. Middle: Hannes observations from the \textit{eye-in-hand} camera. Bottom: Encoder readings for hand opening/closing and wrist flexion/extension, serving as key proprioceptive conditioning. (\textbf{b}) Executed action sequence, illustrating the complex control dynamics: combined wrist pronation and flexion (positive \textbf{Wrist P/S} velocity and decreasing \textbf{Wrist F/E} position) followed by fingers closing (increasing \textbf{Hand O/C} position). The shaded blue area is the total rotation displacement performed by the wrist.}
    \label{fig:test_table_grasp}
\end{figure*}

\subsection{HannesImitationPolicy Deployment}
\label{sec:policy_deployment}
We deploy the \textit{HannesImitationPolicy} on the prosthetic hand. We test the policy for all objects and scenarios, performing 10 trials per object and monitoring the success rate. We consider a task successful if the user approaches, grasps, and lifts the object within 10 seconds. Tab.~\ref{tab:test_results_YCB_train_1_4_7} reports the number of successful trials for each object and summarizes the success rates averaged across each task and all the experiments. Out of a total of 150 trials per scenario, the policy achieves a success rate of $80.6\%$ in the \textit{Table Grasp}, $68\%$ in the \textit{Shelf Grasp}, and $89.3\%$ in the \textit{Human-to-Hannes Handover}. Overall, the \textit{HannesImitationPolicy} obtaines a success rate of $79.3\%$ across 450 trials.

\subsubsection{Table Grasp}
In this scenario, the policy achieves an average success rate of $80.6\%$ (Tab.~\ref{tab:test_results_YCB_train_1_4_7}). Despite the overall high success rate, we observed some failure cases. For instance, the low success rate of the \texttt{044\_flat\_screwdriver} (1/10 trials) is due to the intrinsic difficulty of executing a top-down power grasp on slender shapes, while we could attribute low performance on the \texttt{065-b\_cups} (4/10 trials) to the small object size. The policy obtains high success rates on the other objects.
Fig.~\ref{fig:test_table_grasp} illustrates the execution of the \textit{HannesImitationPolicy}, where the prosthesis successfully grasps the \texttt{004\_sugar\_box} during the \textit{Table Grasp} task.  
Proprioceptive observations from encoder measurements also inform the control policy (Fig.~\ref{fig:test_table_grasp_observations}). The executed action sequences further detail the policy’s smooth and adaptive grasping behavior (Fig.~\ref{fig:test_table_grasp_actions}). Indeed, it is clear to see the combined flexion and pronation of the wrist during the approach phase, finalized by the hand closing to grasp the object.

\subsubsection{Shelf Grasp}
\begin{figure*}[tp]
\vspace{2mm}
\centering
    \subfloat[]{\includegraphics[width=1\columnwidth]{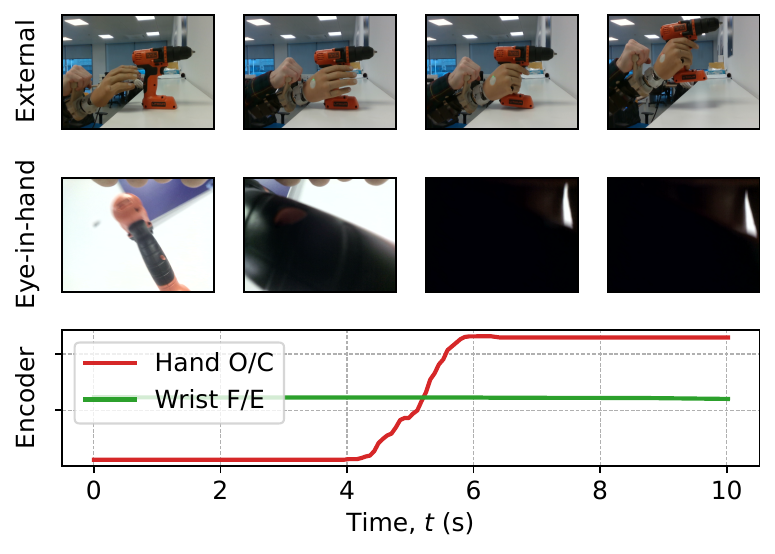}\label{fig:test_shelf_grasp_observations}}
    \subfloat[]{\includegraphics[width=1\columnwidth]{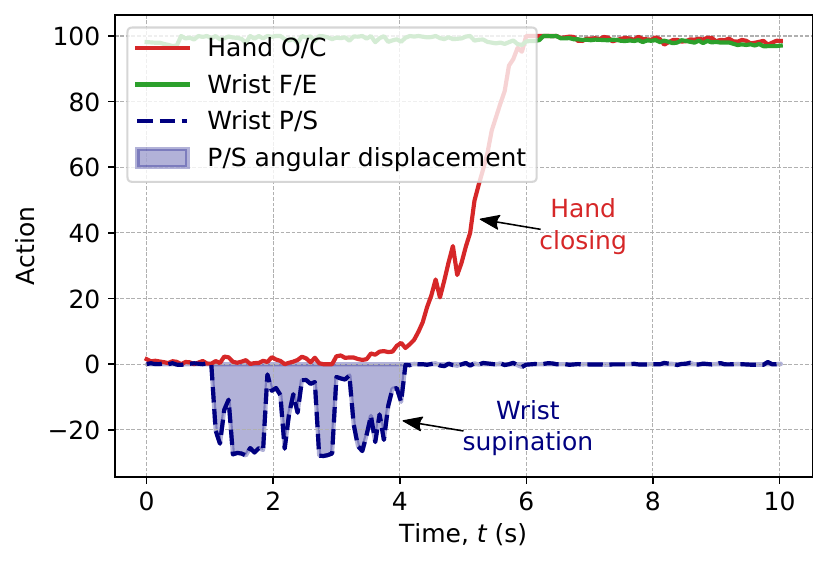}\label{fig:test_shelf_grasp_actions}}
    \caption{\textit{Shelf Grasp} (\textbf{\#2}). \textit{HannesImitationPolicy} deployed on the prosthetic hand to grasp the \texttt{035\_power\_drill} object. (\textbf{a}) External camera capturing the outward wrist rotation required to align the palm with the object handle, observations from the \textit{eye-in-hand} camera embedded into the prosthesis' palm, and encoder readings for hand opening/closing and wrist flexion/extension. (\textbf{b}) Executed action sequences highlighting the wrist supination (negative \textbf{Wrist P/S} velocity) followed by the hand closure (increasing \textbf{Hand O/C} position).}
    \label{fig:test_shelf_grasp}
\end{figure*}

In this task, the policy achieves an average success rate of $68\%$ (Tab.~\ref{tab:test_results_YCB_train_1_4_7}). We can attribute the reduced performance compared to task \textbf{\#1} to the more challenging visual perspective and positioning. For instance, the lowest performance occurs for the \texttt{065-g\_cups} (2/10 trials)---the widest object grasped from the top with wrist flexion and pronation. The policy, nonetheless, exhibits qualitatively meaningful behaviors.
Fig.~\ref{fig:test_shelf_grasp} shows an exemplar trial on the \textit{Shelf Grasp} (\textbf{\#2}) with an object with a proper handle, the \texttt{035\_power\_drill}. The trial begins with the palm facing down. The policy successfully supinates the wrist (outward rotation) and closes the fingers on the object handle. The grasp remains stable while lifting the heavy object (Fig.~\ref{fig:test_shelf_grasp_observations}). Notably, the executed action sequence shows that the fingers begin to close while the wrist is still finishing the outward rotation, akin to human-like grasps (Fig.~\ref{fig:test_shelf_grasp_actions}).

\subsubsection{Human-to-Hannes Handover}
\begin{figure*}[tp]
\vspace{2mm}
\centering
     \subfloat[]{\includegraphics[width=1\columnwidth]{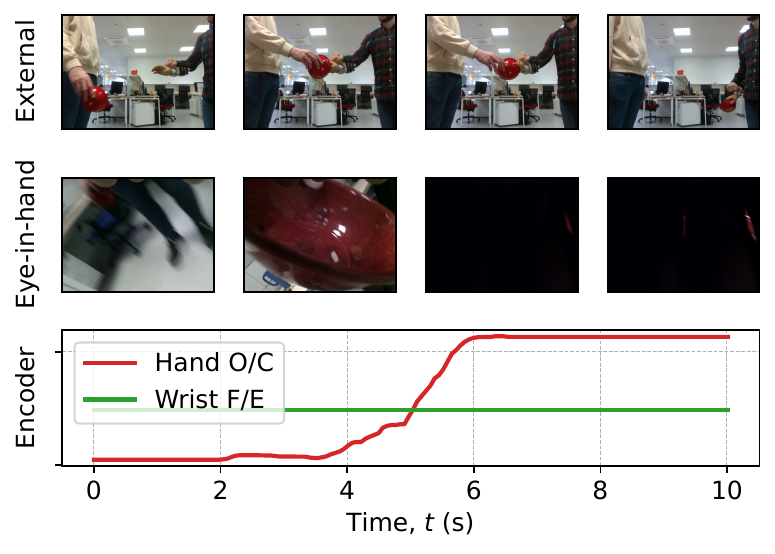}\label{fig:test_handover_observations}}
    \subfloat[]{\includegraphics[width=1\columnwidth]{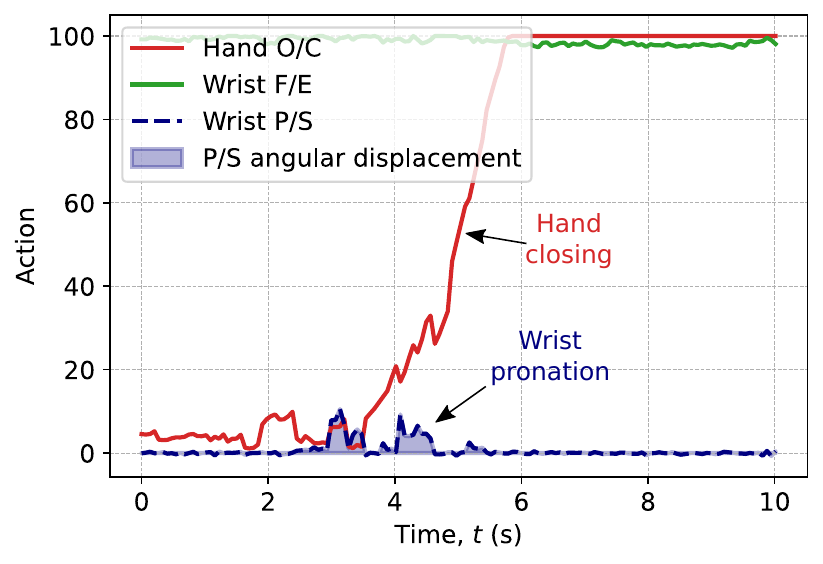}\label{fig:test_handover_actions}}
    \caption{
\textit{Human-to-Hannes Handover} (\textbf{\#3}). \textit{HannesImitationPolicy} deployed on the physical robot for the handover of the \texttt{024\_bowl} object. (\textbf{a}) External view showing the collaboration between the Hannes user and the subject, \textit{eye-in-hand} camera observations from the Hannes palm capturing an unstructured scenario, and position encoder readings. 
(\textbf{b}) Executed action sequences combining minor wrist pronation (positive \textbf{Wrist P/S} velocity) and hand closure (rising \textbf{Hand O/C} position).
    }
    \label{fig:test_human_to_hannes_handover}
\end{figure*}

We conduct the human-to-prosthesis handover experiments with the participation of five subjects transferring the object. In this task, the policy achieves an average of $89.3\%$ success rate (Tab.~\ref{tab:test_results_YCB_train_1_4_7}). We attribute the higher grasping performance with respect to scenario \textbf{\#1} and \textbf{\#2} to the collaboration that naturally emerges between the user and the subject. The policy complements the user-subject interaction during the handover phase. 
For instance, in the \textit{Human-to-Hannes Handover} scenario, the policy satisfactorily grasps the small \texttt{065-b\_cups} (9/10 trials), compared to the \textit{Table Grasp} (4/10 trials) and \textit{Shelf Grasp} (3/10 trials).
Fig.~\ref{fig:test_human_to_hannes_handover} shows an exemplar sequence during a handover experiment, where the subject hands over a slightly tilted \texttt{024\_bowl} to the user guiding the prosthesis, and the policy grasps the object rotating the wrist.

\begin{table}[tp]
\vspace{2mm}
\caption{Test results of \textit{HannesImitationPolicy} deployed on the physical prosthesis. We report task success rates on 3 scenarios for 5 unseen YCB objects.}
\label{tab:test_results_YCB_unseen}
\resizebox{\columnwidth}{!}{%
\begin{tabular}{@{}l|c|c|c@{}}
\toprule
 &
  \begin{tabular}[c]{@{}c@{}}\textbf{Table}\\ \textbf{Grasp}\end{tabular} &
  \begin{tabular}[c]{@{}c@{}}\textbf{Shelf}\\ \textbf{Grasp}\end{tabular} &
  \begin{tabular}[c]{@{}c@{}}\textbf{Human-to-Hannes}\\ \textbf{Handover}\end{tabular} \\ 
  \midrule
\textbf{Unseen YCB Objects} & \textbf{\#1} & \textbf{\#2} & \textbf{\#3} \\
\midrule
\texttt{006\_mustard\_bottle} & $6/10$   &  $6/10$  &  $10/10$   \\
\texttt{010\_potted\_meat\_can}   & $4/10$   &  $8/10$   & $4/10$ \\
\texttt{011\_banana}             & $8/10$   &  $8/10$   &  $10/10$   \\
\texttt{025\_mug}            & $6/10$   &  $6/10$  &    $9/10$   \\
\texttt{055\_baseball}             & $10/10$  & $9/10$   &   $10/10$ \\
\midrule
\multicolumn{1}{l}{\textbf{Aggregate success rate}} & 
\multicolumn{1}{c}{$68\%$} & 
\multicolumn{1}{c}{$74\%$} &
\multicolumn{1}{c}{$86\%$} \\
\midrule
\multicolumn{4}{l}{\textbf{Overall success rate:} $76\%$} \\
\bottomrule
\end{tabular}
}
\end{table}

\subsection{Generalization to Unseen Objects}
\label{sec:generalization_unseen_objects}
We evaluate the \textit{HannesImitationPolicy} on five YCB objects unseen during training across three scenarios without additional fine-tuning. As before, we conduct $10$ trials per object per scenario, totaling $150$ trials. As summarized in Tab.~\ref{tab:test_results_YCB_unseen}, the policy achieves an overall success rate of $76\%$ across the three scenarios ($68\%$, $74\%$, and $86\%$, respectively). Notably, its performance on unseen objects remains consistent with results from the \textit{HannesImitationDataset} (Tab.~\ref{tab:test_results_YCB_train_1_4_7}), demonstrating strong generalization. Moreover, the policy successfully replicates the wrist pronation/supination motions required for grasping objects of different shapes (e.g., spherical, bottle, box). This good generalization capability can be attributed to the diverse object demonstrations in the \textit{HannesImitationDataset} and the expressiveness of diffusion models.

\begin{table}[tp]
\vspace{2mm}
\caption{Comparison of \textit{HannesImitationPolicy} with a state-of-the-art visual-servo wrist controller. On average, the proposed method outperforms \cite{vasile2025continuous} by $13.8\%$.}
\label{tab:comparison}
\resizebox{\columnwidth}{!}{%
\begin{tabular}{@{}l|c|c|c@{}}
\toprule
 &
  \begin{tabular}[c]{@{}c@{}}\textbf{Table}\\ \textbf{Grasp}\end{tabular} &
  \begin{tabular}[c]{@{}c@{}}\textbf{Shelf}\\ \textbf{Grasp}\end{tabular} &
  \begin{tabular}[c]{@{}c@{}}\textbf{Human-to-Hannes}\\ \textbf{Handover}\end{tabular}\\ 
  \midrule
\textbf{Method} & \textbf{\#1} & \textbf{\#2} & \textbf{\#3}\\
\midrule
\textit{HannesImitationPolicy} & $68\%$   &  $74\%$  &  $\textbf{86}\%$   \\
Visual Servoing \cite{vasile2025continuous} & $\textbf{92}\%$   &  $\textbf{80}\%$   &  $16\%$   \\
\bottomrule
\end{tabular}
}
\end{table}

\subsection{Comparison with Visual Servoing Wrist Controller}
\label{sec:comparison_visual_servo}
Finally, we compare the proposed \textit{HannesImitationPolicy} with the visual servoing system for continuous wrist control presented in~\cite{vasile2025continuous}. This framework is based on a segmentation network to detect the object of interest and a visual servoing control scheme to continuously adjust the wrist orientation during the approach. When the user is ready to grasp, they trigger a prediction using EMG signals, the visual servoing control stops running, and the segmentation model predicts the final wrist configuration as either \textit{top} or \textit{side} grasp. Finally, the user completes the grasp by closing the fingers around the object through EMG control. For this method, we consider a grasp successful if (i) the segmentation model correctly identifies and tracks the target object and (ii) the predicted final wrist configuration is correct.

The evaluation is conducted on five YCB objects that are not used for training the policy, with 10 trials per object per scenario (totaling 150 trials). In contrast, we emphasize that the segmentation model for visual servo control was trained on these five objects in uncluttered tabletop scenarios~\cite{vasile2025continuous}.

\textit{HannesImitationPolicy} achieves $76\%$ average success rate, outperforming the visual servo controller ($62.6\%$). As shown in Tab.~\ref{tab:comparison}, the method in~\cite{vasile2025continuous} performs better on the \textit{Table Grasp} and \textit{Shelf Grasp}. This can be explained by the fact that this method was trained on a dataset that included these objects. Additionally, we remark that the visual servo in~\cite{vasile2025continuous} drives the wrist and leaves the fingers' control to the user, whereas our method controls both joints. Conversely, in the \textit{Human-to-Hannes Handover}, our method significantly outperformed the visual servo. In this challenging scenario, the segmentation model in~\cite{vasile2025continuous} fails to distinguish between the object and the subject's body, leading to wrong wrist movements. Overall, \textit{HannesImitationPolicy} demonstrates greater robustness in grasping new objects, whereas the visual servo controller exhibits a performance drop in an unseen scenario.

These results demonstrate the effectiveness of \ac{IL}-based prosthetic control leveraging demonstrations in the \textit{HannesImitationDataset} to enhance generalization across scenarios.


\section{Conclusion \& Future Work}
\label{sec:conclusion}
\acresetall
This paper introduced \textit{HannesImitation}, a control framework based on \ac{IL} for prosthetic hands with an \textit{eye-in-hand} camera. We presented \textit{HannesImitationPolicy}, a novel application of \ac{DP} for learning grasping and human-to-prosthesis handover tasks using the Hannes prosthetic hand with a controllable wrist. Additionally, we released \textit{HannesImitationDataset}, a collection of grasping demonstrations for vision-based \ac{IL} in unstructured environments.
By leveraging the generative capabilities of \ac{DP}, our approach achieved real-time grasping across diverse objects and scenarios, attaining a success rate of $79.3\%$ on in-distribution objects and $76\%$ on unseen objects. 
These findings demonstrate the potential of \ac{IL} for prosthetic control. To the best of our knowledge, this is the first dataset supporting studies in \ac{IL} for the control of prosthetic hands. Future work will focus on expanding our work to tackle more complex actions and environments, comparing with a broader set of baselines, and conducting user studies to assess real-world usability and impact.


\section*{Acknowledgement}
We acknowledge financial support from the PNRR MUR project PE0000013-FAIR, the European Union’s Horizon-JU-SNS-2022 Research and Innovation Programme under the project TrialsNet (Grant Agreement No. 101095871), the project RAISE (Robotics and AI for Socio-economic Empowerment) implemented under the National Recovery and Resilience Plan, Mission 4 funded by the European Union – NextGenerationEU, and the Istituto Nazionale Assicurazione Infortuni sul Lavoro, under grant agreement PR19-PAS-P1, iHannes. 

\bibliography{bib}

\begin{thebibliography}{10}
\providecommand{\url}[1]{#1}
\csname url@samestyle\endcsname
\providecommand{\newblock}{\relax}
\providecommand{\bibinfo}[2]{#2}
\providecommand{\BIBentrySTDinterwordspacing}{\spaceskip=0pt\relax}
\providecommand{\BIBentryALTinterwordstretchfactor}{4}
\providecommand{\BIBentryALTinterwordspacing}{\spaceskip=\fontdimen2\font plus
\BIBentryALTinterwordstretchfactor\fontdimen3\font minus \fontdimen4\font\relax}
\providecommand{\BIBforeignlanguage}[2]{{%
\expandafter\ifx\csname l@#1\endcsname\relax
\typeout{** WARNING: IEEEtran.bst: No hyphenation pattern has been}%
\typeout{** loaded for the language `#1'. Using the pattern for}%
\typeout{** the default language instead.}%
\else
\language=\csname l@#1\endcsname
\fi
#2}}
\providecommand{\BIBdecl}{\relax}
\BIBdecl

\bibitem{ibarz2021train}
J.~Ibarz, J.~Tan, C.~Finn, M.~Kalakrishnan, P.~Pastor, and S.~Levine, ``How to train your robot with deep reinforcement learning: lessons we have learned,'' \emph{The International Journal of Robotics Research}, vol.~40, no. 4-5, pp. 698--721, 2021.

\bibitem{falotico2024learning}
E.~Falotico, E.~Donato, C.~Alessi, E.~Setti, M.~S. Nazeer, C.~Agabiti, D.~Caradonna, D.~Bianchi, F.~Piqu{\'e}, Y.~T. Ansari \emph{et~al.}, ``Learning controllers for continuum soft manipulators: Impact of modeling and looming challenges,'' \emph{Advanced Intelligent Systems}, p. 2400344, 2024.

\bibitem{trent2020narrative}
L.~Trent, M.~Intintoli, P.~Prigge, C.~Bollinger, L.~S. Walters, D.~Conyers, J.~Miguelez, and T.~Ryan, ``A narrative review: current upper limb prosthetic options and design,'' \emph{Disability and Rehabilitation: Assistive Technology}, 2020.

\bibitem{laffranchi2020hannes}
M.~Laffranchi, N.~Boccardo, S.~Traverso, L.~Lombardi, M.~Canepa, A.~Lince, M.~Semprini, J.~A. Saglia, A.~Naceri, R.~Sacchetti \emph{et~al.}, ``The hannes hand prosthesis replicates the key biological properties of the human hand,'' \emph{Science robotics}, vol.~5, no.~46, p. eabb0467, 2020.

\bibitem{oskoei2007myoelectric}
M.~A. Oskoei and H.~Hu, ``Myoelectric control systems—a survey,'' \emph{Biomedical signal processing and control}, vol.~2, no.~4, pp. 275--294, 2007.

\bibitem{chen2023review}
Z.~Chen, H.~Min, D.~Wang, Z.~Xia, F.~Sun, and B.~Fang, ``A review of myoelectric control for prosthetic hand manipulation,'' \emph{Biomimetics}, vol.~8, no.~3, p. 328, 2023.

\bibitem{amsuess2014extending}
S.~Amsuess, P.~Goebel, B.~Graimann, and D.~Farina, ``Extending mode switching to multiple degrees of freedom in hand prosthesis control is not efficient,'' in \emph{2014 36th Annual International Conference of the IEEE Engineering in Medicine and Biology Society}.\hskip 1em plus 0.5em minus 0.4em\relax IEEE, 2014, pp. 658--661.

\bibitem{biddiss2007upper}
E.~Biddiss and T.~Chau, ``Upper-limb prosthetics: critical factors in device abandonment,'' \emph{American journal of physical medicine \& rehabilitation}, vol.~86, no.~12, pp. 977--987, 2007.

\bibitem{di2021hannes}
D.~Di~Domenico, A.~Marinelli, N.~Boccardo, M.~Semprini, L.~Lombardi, M.~Canepa, S.~Stedman, A.~D. Bellingegni, M.~Chiappalone, E.~Gruppioni \emph{et~al.}, ``Hannes prosthesis control based on regression machine learning algorithms,'' in \emph{2021 IEEE/RSJ International Conference on Intelligent Robots and Systems (IROS)}.\hskip 1em plus 0.5em minus 0.4em\relax IEEE, 2021, pp. 5997--6002.

\bibitem{egle_preliminary_2023}
F.~Egle, D.~Di~Domenico, A.~Marinelli, N.~Boccardo, M.~Canepa, M.~Laffranchi, L.~De~Michieli, and C.~Castellini, ``Preliminary {Assessment} of {Two} {Simultaneous} and {Proportional} {Myocontrol} {Methods} for 3-{DoFs} {Prostheses} {Using} {Incremental} {Learning},'' in \emph{2023 {International} {Conference} on {Rehabilitation} {Robotics} ({ICORR})}.\hskip 1em plus 0.5em minus 0.4em\relax IEEE, 2023, pp. 1--6.

\bibitem{scheme2011electromyogram}
E.~Scheme and K.~Englehart, ``Electromyogram pattern recognition for control of powered upper-limb prostheses: state of the art and challenges for clinical use.'' \emph{Journal of Rehabilitation Research \& Development}, vol.~48, no.~6, 2011.

\bibitem{kyranou2018causes}
I.~Kyranou, S.~Vijayakumar, and M.~S. Erden, ``Causes of performance degradation in non-invasive electromyographic pattern recognition in upper limb prostheses,'' \emph{Frontiers in neurorobotics}, vol.~12, p.~58, 2018.

\bibitem{marinelli2023active}
A.~Marinelli, N.~Boccardo, F.~Tessari, D.~Di~Domenico, G.~Caserta, M.~Canepa, G.~Gini, G.~Barresi, M.~Laffranchi, L.~De~Michieli \emph{et~al.}, ``Active upper limb prostheses: A review on current state and upcoming breakthroughs,'' \emph{Progress in Biomedical Engineering}, vol.~5, no.~1, p. 012001, 2023.

\bibitem{zhuang2019shared}
K.~Z. Zhuang, N.~Sommer, V.~Mendez, S.~Aryan, E.~Formento, E.~D’Anna, F.~Artoni, F.~Petrini, G.~Granata, G.~Cannaviello \emph{et~al.}, ``Shared human--robot proportional control of a dexterous myoelectric prosthesis,'' \emph{Nature Machine Intelligence}, vol.~1, no.~9, pp. 400--411, 2019.

\bibitem{gardner2020multimodal}
M.~Gardner, C.~S. Mancero~Castillo, S.~Wilson, D.~Farina, E.~Burdet, B.~C. Khoo, S.~F. Atashzar, and R.~Vaidyanathan, ``A multimodal intention detection sensor suite for shared autonomy of upper-limb robotic prostheses,'' \emph{Sensors}, vol.~20, no.~21, p. 6097, 2020.

\bibitem{guo2023toward}
W.~Guo, W.~Xu, Y.~Zhao, X.~Shi, X.~Sheng, and X.~Zhu, ``Toward human-in-the-loop shared control for upper-limb prostheses: a systematic analysis of state-of-the-art technologies,'' \emph{IEEE transactions on Medical Robotics and Bionics}, vol.~5, no.~3, pp. 563--579, 2023.

\bibitem{starke2022semi}
J.~Starke, P.~Weiner, M.~Crell, and T.~Asfour, ``Semi-autonomous control of prosthetic hands based on multimodal sensing, human grasp demonstration and user intention,'' \emph{Robotics and Autonomous Systems}, vol. 154, p. 104123, 2022.

\bibitem{castro2022continuous}
M.~N. Castro and S.~Dosen, ``Continuous semi-autonomous prosthesis control using a depth sensor on the hand,'' \emph{Frontiers in Neurorobotics}, vol.~16, p. 814973, 2022.

\bibitem{vasile2025continuous}
F.~Vasile, E.~Maiettini, G.~Pasquale, N.~Boccardo, and L.~Natale, ``Continuous wrist control on the hannes prosthesis: a vision-based shared autonomy framework,'' \emph{arXiv preprint arXiv:2502.17265}, 2025.

\bibitem{chi2023diffusionpolicy}
C.~Chi, S.~Feng, Y.~Du, Z.~Xu, E.~Cousineau, B.~Burchfiel, and S.~Song, ``Diffusion policy: Visuomotor policy learning via action diffusion,'' in \emph{Proceedings of Robotics: Science and Systems (RSS)}, 2023.

\bibitem{zitkovich2023rt}
B.~Zitkovich, T.~Yu, S.~Xu, P.~Xu, T.~Xiao, F.~Xia, J.~Wu, P.~Wohlhart, S.~Welker, A.~Wahid \emph{et~al.}, ``Rt-2: Vision-language-action models transfer web knowledge to robotic control,'' in \emph{Conference on Robot Learning}.\hskip 1em plus 0.5em minus 0.4em\relax PMLR, 2023, pp. 2165--2183.

\bibitem{oxe}
O.~X.-E. Collaboration, ``Open x-embodiment: Robotic learning datasets and rt-x models,'' in \emph{2024 IEEE International Conference on Robotics and Automation (ICRA)}, 2024, pp. 6892--6903.

\bibitem{kim2024openvla}
M.~J. Kim, K.~Pertsch, S.~Karamcheti, T.~Xiao, A.~Balakrishna, S.~Nair, R.~Rafailov, E.~Foster, G.~Lam, P.~Sanketi \emph{et~al.}, ``Open{VLA}: An open-source vision-language-action model,'' \emph{arXiv preprint arXiv:2406.09246}, 2024.

\bibitem{black2024pi_0}
K.~Black, N.~Brown, D.~Driess, A.~Esmail, M.~Equi, C.~Finn, N.~Fusai, L.~Groom, K.~Hausman, B.~Ichter \emph{et~al.}, ``{$\pi_0 $}: A vision-language-action flow model for general robot control,'' \emph{arXiv preprint arXiv:2410.24164}, 2024.

\bibitem{zhang2024affordance}
F.~Zhang and M.~Gienger, ``Affordance-based robot manipulation with flow matching,'' \emph{arXiv preprint arXiv:2409.01083}, 2024.

\bibitem{ho2020denoising}
J.~Ho, A.~Jain, and P.~Abbeel, ``Denoising diffusion probabilistic models,'' \emph{Advances in neural information processing systems}, vol.~33, pp. 6840--6851, 2020.

\bibitem{zhang2018deep}
T.~Zhang, Z.~McCarthy, O.~Jow, D.~Lee, X.~Chen, K.~Goldberg, and P.~Abbeel, ``Deep imitation learning for complex manipulation tasks from virtual reality teleoperation,'' in \emph{2018 IEEE international conference on robotics and automation (ICRA)}.\hskip 1em plus 0.5em minus 0.4em\relax Ieee, 2018, pp. 5628--5635.

\bibitem{shafiullah2023bringing}
N.~M.~M. Shafiullah, A.~Rai, H.~Etukuru, Y.~Liu, I.~Misra, S.~Chintala, and L.~Pinto, ``On bringing robots home,'' \emph{arXiv preprint arXiv:2311.16098}, 2023.

\bibitem{chi2024universal}
C.~Chi, Z.~Xu, C.~Pan, E.~Cousineau, B.~Burchfiel, S.~Feng, R.~Tedrake, and S.~Song, ``Universal manipulation interface: In-the-wild robot teaching without in-the-wild robots,'' \emph{arXiv preprint arXiv:2402.10329}, 2024.

\bibitem{dovsen2010cognitive}
S.~Do{\v{s}}en, C.~Cipriani, M.~Kosti{\'c}, M.~Controzzi, M.~C. Carrozza, and D.~B. Popovi{\'c}, ``Cognitive vision system for control of dexterous prosthetic hands: experimental evaluation,'' \emph{Journal of neuroengineering and rehabilitation}, vol.~7, pp. 1--14, 2010.

\bibitem{markovic2015sensor}
M.~Markovic, S.~Dosen, D.~Popovic, B.~Graimann, and D.~Farina, ``Sensor fusion and computer vision for context-aware control of a multi degree-of-freedom prosthesis,'' \emph{Journal of neural engineering}, vol.~12, no.~6, p. 066022, 2015.

\bibitem{cirelli2023semiautonomous}
G.~Cirelli, C.~Tamantini, L.~P. Cordella, and F.~Cordella, ``A semiautonomous control strategy based on computer vision for a hand--wrist prosthesis,'' \emph{Robotics}, vol.~12, no.~6, p. 152, 2023.

\bibitem{10235885}
E.~Ragusa, S.~Dosen, R.~Zunino, and P.~Gastaldo, ``Affordance segmentation using tiny networks for sensing systems in wearable robotic devices,'' \emph{IEEE Sensors Journal}, vol.~23, no.~19, pp. 23\,916--23\,926, 2023.

\bibitem{vasile2022grasp}
F.~Vasile, E.~Maiettini, G.~Pasquale, A.~Florio, N.~Boccardo, and L.~Natale, ``Grasp pre-shape selection by synthetic training: Eye-in-hand shared control on the hannes prosthesis,'' in \emph{2022 IEEE/RSJ International Conference on Intelligent Robots and Systems (IROS)}.\hskip 1em plus 0.5em minus 0.4em\relax IEEE, 2022, pp. 13\,112--13\,119.

\bibitem{10731261}
N.~Kleer, O.~Keil, M.~Feick, A.~Gomaa, T.~Schwartz, and M.~Feld, ``Incorporation of the intended task into a vision-based grasp type predictor for multi-fingered robotic grasping,'' in \emph{2024 33rd IEEE International Conference on Robot and Human Interactive Communication (ROMAN)}, 2024, pp. 1301--1307.

\bibitem{zandigohar2019towards}
M.~Zandigohar, M.~Han, D.~Erdo{\u{g}}mu{\c{s}}, and G.~Schirner, ``Towards creating a deployable grasp type probability estimator for a prosthetic hand,'' in \emph{International Workshop on Design, Modeling, and Evaluation of Cyber Physical Systems}.\hskip 1em plus 0.5em minus 0.4em\relax Springer, 2019, pp. 44--58.

\bibitem{10375204}
F.~Hundhausen, S.~Hubschneider, and T.~Asfour, ``Grasping with humanoid hands based on in-hand vision and hardware-accelerated cnns,'' in \emph{2023 IEEE-RAS 22nd International Conference on Humanoid Robots (Humanoids)}, 2023, pp. 1--7.

\bibitem{10643668}
C.~Peng, D.~Yang, D.~Zhao, M.~Cheng, J.~Dai, and L.~Jiang, ``Viiat-hand: A reach-and-grasp restoration system integrating voice interaction, computer vision, auditory and tactile feedback for non-sighted amputees,'' \emph{IEEE Robotics and Automation Letters}, vol.~9, no.~10, pp. 8674--8681, 2024.

\bibitem{zandigohar2024multimodal}
M.~Zandigohar, M.~Han, M.~Sharif, S.~Y. G{\"u}nay, M.~P. Furmanek, M.~Yarossi, P.~Bonato, C.~Onal, T.~Pad{\i}r, D.~Erdo{\u{g}}mu{\c{s}} \emph{et~al.}, ``Multimodal fusion of emg and vision for human grasp intent inference in prosthetic hand control,'' \emph{Frontiers in Robotics and AI}, vol.~11, p. 1312554, 2024.

\bibitem{stracquadanio2025bring}
G.~Stracquadanio, F.~Vasile, E.~Maiettini, N.~Boccardo, and L.~Natale, ``Bring your own grasp generator: Leveraging robot grasp generation for prosthetic grasping,'' \emph{arXiv preprint arXiv:2503.00466}, 2025.

\bibitem{boccardo2023development}
N.~Boccardo, M.~Canepa, S.~Stedman, L.~Lombardi, A.~Marinelli, D.~Di~Domenico, R.~Galviati, E.~Gruppioni, L.~De~Michieli, and M.~Laffranchi, ``Development of a 2-dofs actuated wrist for enhancing the dexterity of myoelectric hands,'' \emph{IEEE Transactions on Medical Robotics and Bionics}, vol.~6, no.~1, pp. 257--270, 2023.

\bibitem{calli2015benchmarking}
B.~Calli, A.~Walsman, A.~Singh, S.~Srinivasa, P.~Abbeel, and A.~M. Dollar, ``Benchmarking in manipulation research: Using the yale-cmu-berkeley object and model set,'' \emph{IEEE Robotics \& Automation Magazine}, vol.~22, no.~3, pp. 36--52, 2015.

\bibitem{he2016deep}
K.~He, X.~Zhang, S.~Ren, and J.~Sun, ``Deep residual learning for image recognition,'' in \emph{Proceedings of the IEEE conference on computer vision and pattern recognition}, 2016, pp. 770--778.

\end{thebibliography}
\bibliographystyle{IEEEtran}

\end{document}